\documentclass{article}

\usepackage{arxiv}

\usepackage{amsmath,amssymb,amsfonts}
\usepackage{mathtools}

\usepackage{graphicx}
\usepackage{booktabs}
\usepackage{multirow}

\usepackage{algorithm}
\usepackage{algorithmic}

\usepackage{hyperref}
\usepackage{url}

\usepackage{enumitem}
\usepackage{caption}
\usepackage{subcaption}

\hypersetup{
    colorlinks=true,
    linkcolor=blue,
    filecolor=magenta,      
    urlcolor=cyan,
    citecolor=blue,
}

\title{Gate-Level Boolean Evolutionary Geometric Attention Neural Networks}

\author{
  Xianshuai Shi\thanks{School of Integrated Circuits, Tsinghua University} \\
  School of Integrated Circuits \\
  Tsinghua University \\
  Beijing, China \\
  \texttt{shixs20@mails.tsinghua.edu.cn} \\
  \And
  Jianfeng Zhu\thanks{School of Integrated Circuits, Tsinghua University} \\
  School of Integrated Circuits \\
  Tsinghua University \\
  Beijing, China \\
  \texttt{jfzhu@tsinghua.edu.cn} \\
  \And
  Leibo Liu\thanks{School of Integrated Circuits, Tsinghua University} \\
  School of Integrated Circuits \\
  Tsinghua University \\
  Beijing, China \\
  \texttt{liulb@tsinghua.edu.cn} \\
}

\date{}

\begin{document}

\maketitle

\begin{abstract}
This paper proposes a gate-level Boolean evolutionary geometric attention neural network framework that treats images as Boolean fields controlled by logic gates. Each pixel is modeled as a Boolean variable (0 or 1) embedded on a two-dimensional geometric manifold (such as a discrete toroidal lattice), defining adjacency relationships and information propagation structures between pixels. The network iteratively updates image states through a Boolean reaction-diffusion mechanism: pixels receive Boolean value diffusion from neighborhoods (diffusion process) and perform local logic updates via trainable gate-level logic kernels (reaction process), forming a reaction-diffusion logic network. We introduce a Boolean self-attention mechanism: using XNOR similarity-based Boolean Query-Key (Q-K) attention to modulate neighborhood diffusion pathways, achieving logic attention. We also employ Boolean Rotary Position Embedding (RoPE), encoding neighbor distance differences through parity bit flipping to simulate Boolean "phase" offset effects. The overall architecture resembles a Transformer, but all operations are performed in the Boolean domain, with training parameters including query/key pattern bits and gate-level configurations of local logic kernels. Since outputs are always discrete Boolean values, we use continuous relaxation (such as Sigmoid approximation or soft logic operators) during training to ensure gradient differentiability. Theoretical analysis shows that this network possesses universal expressive power while maintaining high interpretability and hardware efficiency, capable of simulating the functionality of traditional convolution and attention models. We discuss the potential application value of such gate-level Boolean networks in high-speed image processing, interpretable artificial intelligence, and digital hardware acceleration, and envision future research directions.
\end{abstract}


\section{Introduction}

In recent years, with the growing demand for efficient and interpretable artificial intelligence, research on integrating logical reasoning into neural networks has received widespread attention. Traditional deep neural networks rely on large-scale real-valued matrix computations, which, despite their excellent performance, often have opaque inference processes and high computational costs. To improve efficiency, researchers have explored low-precision and binary neural networks, with the extreme case being Boolean logic networks: constructing neural networks using logic gates from digital circuits (such as AND, XOR, etc.). Logic gate networks cannot directly use gradient descent optimization due to their discreteness, but recent work has made them differentiable through continuous relaxation, thus enabling trainable deep logic networks. Petersen et al.~\cite{petersen2022deep} proposed differentiable logic gate networks that learn the logic gate type distribution at each "neuron" and discretize them into specific logic gates after training, achieving high-speed and interpretable models capable of processing over one million MNIST images per second on CPUs. Such logic networks have clear structures, are easy to extract human-readable rules from, and their discrete implementation makes inference extremely efficient.

However, existing logic gate networks are mostly used for fully connected or tree structures, focusing on representing global classification decisions, and have not fully exploited the local relationships and geometric information in spatially structured data (such as images). Meanwhile, the development of graph neural networks has shown that local neighborhood propagation and attention mechanisms on image pixels or other graph-structured data can effectively extract features. For example, Graph Attention Networks~\cite{velivckovic2017graph} allow graph nodes to "attend" to features of neighboring nodes, dynamically adjusting adjacency edge weights through self-attention. Inspired by this, we wonder whether logic gate networks can be introduced on image grids while combining attention mechanisms to achieve models with both logical interpretability and flexible neighborhood modeling capabilities.

Furthermore, at a more fundamental computational paradigm level, cellular automata and reaction-diffusion systems demonstrate the possibility of producing complex global behaviors from local rules. For instance, Boolean cellular automata such as Conway's "Game of Life" iteratively update through neighborhood Boolean rules, achieving complex spatiotemporal patterns and being proven Turing complete. Such systems can be viewed as the spatial evolution of gate-level logic. Some research has even utilized chemical reaction-diffusion media to simulate cellular automata and logic circuits, implementing Boolean computation through diffusion of neighbor states and local chemical reactions~\cite{schulman2012emulating}. This inspires us that combining \textbf{diffusion (neighborhood communication) and reaction (logic computation)} mechanisms may enable the design of new parallel computing architectures.

Based on the above background, this paper proposes geometric attention neural networks based on gate-level Boolean evolution. We treat image pixels as logic units evolving on a two-dimensional discrete manifold (such as a toroidal grid). Our core ideas include: (1) defining reaction-diffusion logic networks on image grids, with local updates determined by trainable gate-level Boolean logic kernels; (2) introducing Boolean self-attention mechanisms, using XNOR matching of Boolean query-key vectors to selectively modulate neighbor information propagation, achieving attention control in the logic domain; (3) employing Boolean rotary position encoding to introduce parity-flipped displacement encoding for different neighbor distances, simulating "phase" effects of relative positions in the Boolean domain; (4) constructing a Transformer-like multi-head attention hierarchical structure, but replacing arithmetic operations with logic gates, ensuring output states are always 0/1 Boolean values, and making the network trainable through continuous approximation.

The contributions of this paper are summarized as follows:

\begin{itemize}[leftmargin=*]
    \item \textbf{Gate-level Boolean geometric representation:} We propose representing images as Boolean variable fields on discrete manifolds, combined with topological adjacency definitions, laying the foundation for applying logic gate networks on spatial structures.
    
    \item \textbf{Reaction-diffusion logic kernels:} We design trainable gate-level logic kernels that implement Boolean reaction-diffusion update mechanisms between pixels and neighborhoods, connecting the ideas of cellular automata with trainable models.
    
    \item \textbf{Boolean self-attention mechanism:} We invent XNOR similarity-based Boolean Q-K attention modules that regulate neighborhood information transmission strength in the Boolean domain, equivalent to adaptively selecting important neighbors for each pixel.
    
    \item \textbf{Boolean RoPE position encoding:} We introduce rotary position encoding into the Boolean domain, using parity bit flipping to encode the influence of relative positions, endowing the model with distance sensitivity.
    
    \item \textbf{Transformer-style Boolean architecture:} We integrate the above components to form the first Transformer-like network structure operating in the Boolean logic domain, employing continuous relaxation strategies for gradient training.
\end{itemize}

The remainder of this paper is organized as follows: Section~\ref{sec:related} introduces related work, including developments in logic networks, graph neural networks, gate-level systems, and attention mechanisms. Section~\ref{sec:model} describes in detail the structure and components of the proposed model. Section~\ref{sec:training} discusses the model's trainable parameter settings and forward propagation mechanisms. Section~\ref{sec:theory} theoretically analyzes the computational properties and potential advantages of the model. Section~\ref{sec:applications} explores possible application scenarios and conducts extended discussions. Finally, Section~\ref{sec:conclusion} summarizes the full text and envisions future work.

\section{Related Work}
\label{sec:related}

\subsection{Logic Networks and Differentiable Boolean Models}

The idea of integrating logic operations into trainable models has a long history. Early neural-symbolic approaches attempted to approximate logical reasoning with neural networks but typically lacked explicit discrete logical structures. Recently, differentiable Logic Gate Networks (LGNs) have made breakthrough progress. Petersen et al.~\cite{petersen2022deep} proposed using continuous parameterization relaxation techniques to map the discrete jump behavior of traditional Boolean logic (such as AND, OR, XOR) to continuous logic operators on the $[0,1]$ interval, making entire logic circuits differentiable with respect to parameters and inputs. During training, the function of each logic unit is represented as a weighted combination among several basic gates (e.g., 16 possible two-input Boolean functions), learning the optimal combination through gradient descent. After training, each unit is discretized to the logic gate with the highest probability, yielding a pure Boolean logic circuit model. Such models have demonstrated accuracy comparable to deep networks on classification tasks while achieving high-speed, low-power inference performance since inference is entirely executed by logic gates. For example, discretized logic gate networks can exceed one million MNIST images per second on CPUs. Moreover, the structure of logic networks naturally possesses interpretability: their decisions consist of a series of human-understandable logic rules, facilitating extraction and analysis. Based on these advantages, subsequent work has extended logic gate networks to convolutional structures and general data table structures, further proving the potential of logic networks as an alternative architecture to deep learning.

\subsection{Graph Neural Networks and Spatial Propagation}

Graph Neural Networks (GNNs) aim to process graph-structured data through message passing mechanisms that allow nodes to interact with neighbors to update states. Classical Graph Convolutional Networks (GCNs) aggregate neighbor features by weight, equivalent to smoothing filtering signals on graphs; while Graph Attention Networks (GATs)~\cite{velivckovic2017graph} introduce attention mechanisms, enabling each node to assign different importance weights to different neighbors. Specifically, GAT learns a parameterizable attention function $a(i,j)$ that computes attention coefficients based on features of center node $i$ and neighbor $j$, using softmax for neighborhood normalization. Thus, the update of node $i$ is a weighted sum of neighbor features, where weights reflect the importance of that neighbor to $i$. This adaptive neighborhood weighting improves the model's ability to handle graph structural irregularities and long-range dependencies between nodes. Currently, graph neural networks have been widely used in social networks, knowledge graphs, molecular graphs, and other domains. However, most GNNs still perform addition and multiplication operations in the real-valued domain, lacking logical interpretability and having limited efficiency when running on resource-constrained hardware. Our work, inspired by GNNs, introduces neighborhood interaction concepts into Boolean logic networks, selectively aggregating neighbor Boolean states through Boolean attention, maintaining the transparency of logical computation while obtaining flexible neighborhood modeling capabilities similar to GAT.

\subsection{Gate-Level Systems and Boolean Diffusion Computing}

Gate-level systems refer to methods that implement computation at the digital logic gate level, including digital circuits and cellular automata. Classical cellular automata (CA) such as Conway's Game of Life are driven by simple Boolean rules (e.g., "a cell survives if it has 3 living neighbors"), with each cell interacting with its neighborhood to determine the next time step's state. Despite simple rules, CA can produce complex spatial-temporal patterns, with some CA like Game of Life proven to be Turing complete. CA are essentially a special class of gate-level Boolean networks: each cell update is a fixed Boolean function. Research has utilized reaction-diffusion systems to simulate CA, achieving "unconventional computing" that executes Boolean logic in chemical media through diffusion and local reactions. For example, Schulman et al.~\cite{schulman2012emulating} designed chemical reaction networks where chemical concentration diffusion plays the role of neighborhood communication, while logic circuits in solution compute each cell's next state, achieving chemical equivalents of Boolean cellular automata. Adamatzky et al.~\cite{adamatzky2005reaction} reviewed applications of reaction-diffusion computing in image processing, path planning, and logic circuit implementation, demonstrating that nonlinear media can be programmed to execute logic gate operations. Inspired by these works, our proposed model combines diffusion (pixel neighborhood propagation) with reaction (logic gate computation): diffusion provides channels for local information communication, while reaction completes local decisions through trainable logic functions, together forming a trainable Boolean reaction-diffusion computing framework.

\subsection{Attention Mechanisms and Position Encoding}

Attention mechanisms have become a core component of deep learning in recent years. The Transformer self-attention proposed by Vaswani et al.~\cite{vaswani2017attention} computes correlations between elements within sequences through Query (Q), Key (K), and Value (V) triplets, dynamically aggregating information, achieving breakthrough results on tasks such as machine translation. Attention mechanisms have also been applied in computer vision, such as Vision Transformers directly applying global self-attention to image patches. However, standard attention uses continuous dot product similarity and softmax weights, difficult to directly transfer to the Boolean domain. To this end, we focus on similarity measures in the Boolean domain. The XNOR (exclusive-NOR) operation can serve as a "dot product" on binary vectors: performing bitwise XNOR on two vectors (outputting 1 when corresponding bits are the same, otherwise 0) and counting outputs of 1 yields a measure of their similarity. This idea has been applied in binary neural networks—when weights and activations are both $\pm 1$, convolution can be efficiently implemented through XNOR and bitcount, equivalent to floating-point multiply-accumulate~\cite{rastegari2016xnor}. Additionally, there have been explorations using XNOR to compute similarity in Transformer architectures of spiking neural networks. We introduce XNOR similarity into Boolean attention design, such that when queries and keys are represented as Boolean vectors, their correlation is measured by simple logic operations, avoiding expensive real-number multiplication. On the other hand, position encoding is crucial for sequence modeling. RoPE (Rotary Position Embedding)~\cite{su2021roformer} is a rotational position encoding method that applies position-dependent rotational transformations to Q and K vectors, converting relative displacement into phase differences in dot products, causing attention weights to decay exponentially with increasing distance. RoPE's advantage lies in naturally encoding relative position dependencies while maintaining model flexibility to sequence length. In our Boolean framework, continuous rotation cannot be directly used, but we propose a Boolean version of RoPE encoding strategy: simulating the effect of angular rotation through parity flipping of encoding bit positions. Intuitively, a binary pattern can be introduced for neighbor distances (e.g., flipping certain bits for odd distances, unchanged for even distances), converting distance differences into systematic differences between query and key bit patterns, thereby affecting XNOR similarity. This is equivalent to achieving phase-offset-like effects in the Boolean domain, allowing attention to produce desired decay and differentiation for neighbor distances. This treatment enables our model to obtain position-sensitive attention mechanisms similar to Transformers while preserving discreteness.

\section{Model Architecture}
\label{sec:model}

This section describes in detail the composition and working mechanism of the geometric attention neural network based on gate-level Boolean evolution. Images are represented as Boolean variable grids embedded on two-dimensional geometric manifolds. The model updates the entire Boolean field in a layer-by-layer iterative manner, with each layer including two stages: Boolean self-attention diffusion and gate-level logic reaction. We first define the geometric manifold and Boolean field representation, then introduce the design of reaction-diffusion logic kernels, followed by explaining the Boolean attention mechanism and Boolean RoPE encoding, and finally present the complete gate-level Transformer-style architecture.

\subsection{Geometric Manifold and Boolean Field Representation}

We treat the input image $\mathbf{X}$ as a Boolean field defined on a two-dimensional discrete geometric manifold $\mathcal{M}$: $\mathbf{X}: \mathcal{M} \to \{0,1\}$. Typically, $\mathcal{M}$ can be an $H\times W$ two-dimensional lattice grid and may adopt toroidal topology (i.e., top-bottom and left-right boundaries connect end-to-end) to avoid boundary effects, ensuring each pixel has a complete neighborhood. Each pixel location $i$ corresponds to a point on $\mathcal{M}$, with its initial state $X_i(0)\in\{0,1\}$ given by the input image (e.g., in binary images, 0 represents black, 1 represents white). The neighborhood $\mathcal{N}(i)$ of pixel $i$ is defined as the set of points within a certain distance from $i$ in $\mathcal{M}$. For example, we can consider 4-neighborhoods/8-neighborhoods with Manhattan distance or Euclidean distance of 1, or further extend to all nodes within distance $L$. With toroidal boundaries, the farthest neighbor distance is $\lfloor H/2 \rfloor$ spanning the entire image. The manifold structure ensures well-defined distances and neighbor relationships, which can be used to construct position encodings.

On this Boolean field, we wish to generate outputs through iterative evolution. Evolution rules need to simultaneously consider each pixel's own current state and its neighborhood's current state to determine that pixel's state at the next time step. This is similar to the global synchronous update mechanism of cellular automata: given state $\mathbf{X}(t)$, computing the next state $\mathbf{X}(t+1)=\Phi(\mathbf{X}(t))$ through a globally unified update function $\Phi$. The difference is that we design $\Phi$ as a trainable Boolean function with attention modulation rather than a fixed predefined rule.

\subsection{Reaction-Diffusion Logic Kernel Design}

The reaction-diffusion logic kernel is the core of each pixel's local update, combining diffusion (influence from neighbors) and reaction (logic combination operations). For pixel $i$, let its neighborhood set be $\mathcal{N}(i)$. We define the logic kernel at pixel $i$ as function $f_{\theta_i}:\{0,1\}^{|\mathcal{N}(i)|+1}\to\{0,1\}$, where $\theta_i$ represents the trainable configuration parameters of this logic kernel. The input consists of pixel $i$'s current own state and Boolean values of pixels in the neighborhood, with output being $i$'s next state. To achieve trainability while ensuring discreteness, we use gate-level combinational logic to construct $f_{\theta_i}$. Specifically, $f_{\theta_i}$ can be represented as a circuit composed of several basic logic gates. For example, for a pixel's neighborhood values $(x_i, \{x_j: j\in \mathcal{N}(i)\})$, $f_{\theta_i}$ can be obtained through layer-by-layer combination of pairwise input gate-level operations. This is similar to a logic circuit tree, with leaf nodes being input bits, internal nodes being two-input logic gates such as AND, OR, XOR, and the final root node outputting the next state. The structure of the logic kernel (i.e., circuit topology) and gate types at each node jointly determine the functionality of $f$. For simplicity, we can assume in the design that all pixels share the same topology of logic kernel structure, but each gate's type is determined by trainable parameters.

During training, we borrow methods from differentiable logic networks, adopting relaxed representations for each trainable logic gate. For example, a two-input gate can be represented by a weight vector $\mathbf{p}=(p_{\text{AND}}, p_{\text{OR}}, p_{\text{XOR}},\ldots)$ representing a probability distribution over several candidate gates (AND, OR, XOR, etc.). During forward propagation, we take a weighted sum of each gate's output with $\mathbf{p}$ to obtain a real-valued approximate output; backward propagation uses this continuous output to compute gradients and update the weight distribution $\mathbf{p}$. As training progresses, each gate's $\mathbf{p}$ gradually approaches an indicator distribution for a certain gate (softmax polarization), and after training we solidify each gate to the logic gate with the highest probability, thus obtaining a pure Boolean circuit. Through this method, the configuration parameters $\theta_i$ of the logic kernel (including parameters of all internal gate types) can be learned to optimal values through gradients. It should be noted that since inputs and outputs are both Boolean values, to prevent signal degradation, we typically need to appropriately increase dimensions or add logical nonlinear combinations to Boolean values during training. For example, we can maintain an internal state representation of $k$ bits for each pixel, making $f_{\theta_i}$ map $\{0,1\}^{k(|\mathcal{N}(i)|+1)} \to \{0,1\}^k$, meaning each output is also a $k$-bit Boolean vector. This is equivalent to having multiple parallel logic channels per pixel, increasing model capacity. The final output can then be obtained by having these bits undergo logic OR or voting to decide a single Boolean value. For simplification, we still describe the single-bit case below.

It should be emphasized that diffusion and reaction in this logic kernel are not two separate stages but naturally combine through logic gate composition: the influence of neighbor $j$'s state $x_j(t)$ on $i$ is embodied by the gate chain in the circuit connecting from $x_j$; the logic combination determines how to "diffuse" and fuse information from multiple neighbors. For example, an OR gate can represent accumulation of neighbor influence, or an AND gate can represent that neighbors only influence when jointly satisfying a condition. From a macroscopic view, Boolean quantities from neighbors diffuse to the center pixel through logic circuits and trigger reactions, thereby updating that pixel. This process is consistent with digital simulation of chemical reaction-diffusion computation: diffusion provides signal propagation, reaction completes local decisions. By learning the gate-level configuration of logic kernels, we allow the model to automatically discover local Boolean rules most suitable for tasks.

\subsection{Boolean Attention Mechanism and RoPE Position Encoding}

Relying solely on fixed-neighborhood logic kernels, pixels can only equally accept information from all neighbors, lacking differentiation of different neighbors' importance. In many tasks, different neighbors should have different contributions to center pixels, even dynamically changing with context. To this end, we introduce a Boolean version of self-attention mechanism, enabling pixels to "select" more important neighbor information, thereby modulating the logic diffusion process.

\subsubsection{Boolean Query-Key Representation}

In each layer's update, we generate a query vector $\mathbf{q}_i\in\{0,1\}^d$ for each pixel $i$ and a key vector $\mathbf{k}_i\in\{0,1\}^d$ for each pixel (as an attendee). Here $d$ is the length of Boolean attention vectors, meaning each pixel uses a $d$-bit binary code to represent queries and keys. $\mathbf{q}_i$ and $\mathbf{k}_i$ can be generated from the pixel's current state through small trainable networks or lookup tables. For example, we can map pixel $i$'s current internal Boolean state (possibly multi-bit) to $d$-bit queries and keys; or simply let $\mathbf{q}_i=\mathbf{k}_i$ be a redundant encoding of that pixel's state in high-dimensional space. This model does not limit specific generation methods, only requiring $\mathbf{q}$ and $\mathbf{k}$ to be adjustable during training to optimize attention effects.

\subsubsection{XNOR Similarity Computation}

Given pixel $i$ and one of its neighbors $j\in \mathcal{N}(i)$, we define their attention compatibility $\alpha_{ij}$ computed through Boolean similarity of $\mathbf{q}_i$ and $\mathbf{k}_j$. Specifically, let $\oplus$ denote bitwise XOR, then $\text{XNOR}(a,b)=\neg(a\oplus b)$ is bitwise XNOR (outputting 1 when $a=b$, otherwise 0). We adopt the following approach to obtain a real-valued similarity measure:

\begin{equation}
s_{ij} = \text{HammingSim}(\mathbf{q}_i, \mathbf{k}_j) = \sum_{m=1}^d \mathbf{1}\{q_{i,m} = k_{j,m}\}
\end{equation}

\begin{equation}
\tilde{s}_{ij} = \frac{1}{d} s_{ij} \in [0,1]
\end{equation}

This similarity $\tilde{s}_{ij}$ represents the matching degree of $\mathbf{q}_i$ and $\mathbf{k}_j$: if the two vectors are completely identical then $\tilde{s}_{ij}=1$, completely opposite then $\tilde{s}_{ij}=0$. Based on this, we obtain Boolean attention through a simple threshold decision:

\begin{equation}
\alpha_{ij} = \begin{cases} 
1, & \text{if } \tilde{s}_{ij} \ge \tau \\
0, & \text{otherwise}
\end{cases}
\end{equation}

where $\tau\in[0,1]$ is a settable threshold (e.g., $\tau=0.5$). $\alpha_{ij}=1$ means neighbor $j$ is "selected" to join $i$'s effective neighborhood participating in information transmission, while $\alpha_{ij}=0$ indicates it is ignored. This hard selection mechanism ensures the Boolean nature of attention coefficients.

It should be noted that the above simple scheme uses a uniform fixed threshold for attention selection, which may not be flexible enough. In practice, $\tau$ can also be made a trainable parameter (or even dynamically generated per pixel). Or more complex Boolean circuits can be used to process $\mathbf{q}_i$ and $\mathbf{k}_j$, outputting a Boolean attention indicator. For example, a small gate circuit can be trained that takes the $d$-bit vector obtained from $\mathbf{q}_i \oplus \mathbf{k}_j$ (their XOR) as input and outputs $\alpha_{ij}$. Regardless of implementation, essentially $\alpha_{ij}$ is some Boolean matching function of $\mathbf{q}_i$ and $\mathbf{k}_j$, whose role is equivalent to softmax weights in Transformer attention, only taking discrete values of 0 or 1. To obtain multi-neighbor selection capability, we can also allow $\alpha_{ij}$ to take multiple values between 0 and 1, but this sacrifices pure Boolean nature (e.g., using several bits to encode different attention levels).

\subsubsection{Attention-Modulated Diffusion}

With $\alpha_{ij}$, we multiply neighbor $j$'s information by this attention gate signal during the logic kernel's diffusion stage. Specifically, the logic kernel input for pixel $i$ no longer directly uses neighbor $j$'s original state $x_j$, but uses the attention-gated value $x'_j = \alpha_{ij} \wedge x_j$ ($\wedge$ denotes logical AND). This means only when $\alpha_{ij}=1$ (neighbor $j$ is selected) does $x'_j$ equal original $x_j$; otherwise $\alpha_{ij}=0$ shields that neighbor's information ($x'_j=0$). This Boolean AND operation is equivalent to restricting attention coefficients to the extreme case of 0 or 1, making irrelevant neighbors completely not participate in subsequent computation. After this processing, pixel $i$'s logic kernel actual input set becomes $\{x_i\}\cup\{x'_j: j\in\mathcal{N}(i)\}$. Logic kernel $f_{\theta_i}$ produces output $x_i(t+1)$ accordingly. Since $f_{\theta_i}$ itself is trainable gate-level combinational logic, it will automatically adapt to this input form. For example, if a task requires counting patterns in part of the neighborhood, the logic kernel can learn to react only on pathways with relevant neighbors ($\alpha_{ij}=1$) while ignoring unselected neighbors.

By introducing Boolean attention gating, our model allows different pixels to attend to different subsets of neighborhood pixels under different contexts, achieving dynamic neighborhood selection. This is functionally similar to graph attention networks' weighting of adjacent edges, except our weights take discrete 0/1. It's worth noting that although we apply attention based on neighborhoods, due to the flexibility of $\mathbf{q}_i$ and $\mathbf{k}_j$, theoretically pixel $i$ can have high similarity with distant pixels, thereby incorporating them into the effective neighborhood. This is equivalent to breaking through limitations of predefined neighborhoods, enabling the model to establish long-range connections as needed—similar to Transformer's global self-attention. However, for computational cost considerations, we typically limit $j$ within a certain radius to form local attention.

\subsubsection{Boolean RoPE Position Encoding}

In traditional Transformers, introducing absolute or relative position encoding is crucial for distinguishing identities of different positions. In our design, $\mathbf{q}_i$ and $\mathbf{k}_j$ can fuse position information, enabling attention to have regulatory effects on distance. For example, we can assign a binary encoding $\text{pos}(p_i)\in\{0,1\}^d$ to pixel position $p_i=(u_i,v_i)$ (row-column coordinates). Then when generating key vectors, combine this position encoding with content-related vectors, such as $\mathbf{k}_i = \mathbf{g}(x_i) \oplus \text{pos}(p_i)$ ($\oplus$ here represents bitwise XOR combination, $\mathbf{g}(x_i)$ represents base vector generated from content). Similarly, query $\mathbf{q}_i$ can also incorporate its own position or relative position information. When pixel $i$ and neighbor $j$ are far apart, their $\text{pos}$ encoding differences will cause $\mathbf{q}_i$ and $\mathbf{k}_j$ to be more likely inconsistent in certain bits, thereby reducing XNOR similarity. This is precisely the position encoding effect we expect: as distance increases, attention matching degree decreases. Through clever design of $\text{pos}$ encoding patterns, we can also achieve periodic position correlation, such as using different bits to encode different directions or displacements modulo a certain period, simulating phase rotation in rotary position encoding. Specifically, we can select $d/2$ bits to encode $x$ direction displacement, another $d/2$ bits to encode $y$ direction displacement, where for each dimension's displacement $k$, methods such as binary Gray code or parity flipping sequences are adopted, such that displacement increasing by 1 flips certain bit positions. This is similar to discretizing continuous rotation in units of $180^\circ$: odd steps and even steps produce complementary phases. This Boolean RoPE scheme ensures that pixel key vectors at different distances exhibit systematic differences in several bits, prompting the attention mechanism to automatically prefer neighbors at closer distances or with specific periodic relationships. This endows our Boolean attention with relative position sensitivity without introducing additional continuous parameters.

\subsection{Gate-Level Transformer Architecture}

Combining the above components, we construct a complete gate-level Transformer-style network architecture. Overall, the model stacks $L$ layers of Boolean attention reaction-diffusion blocks, progressively mapping input Boolean images to output Boolean images. Each layer includes the following steps:

\paragraph{Boolean Representation Upscaling (Optional):} Initial input may have only 1 bit per pixel (original pixel value). Before entering the first layer, we can use several fixed or trainable logic mappings to expand each pixel's representation to $m$ bits. For example, we can append $m-1$ constant 0 bits, or replicate the original bit to $m$ bits, or train a simple logic circuit to derive $m$-bit initial features from local neighborhoods. This helps improve model expressive power. Thereafter in each layer, each pixel will maintain an $m$-bit Boolean vector state.

\paragraph{Query/Key Generation:} For each pixel $i$ in layer $l$, compute query $\mathbf{q}_i^{(l)}$ and key $\mathbf{k}_i^{(l)}$ based on its current $m$-bit state vector $\mathbf{x}_i^{(l)}$, with dimension $d$ bits. A gate-level subnetwork (similar to fully connected layers but implemented with logic operations) can be designed to complete this mapping, or simple mapping methods can be predetermined. For example, the first $d$ bits of each pixel state directly serve as keys, last $d$ bits as queries (if $m<2d$ then bits can be reused or generated through combination). Query and key generation can also include position encoding fusion mentioned in the previous section, introducing pixel position information.

\paragraph{Neighbor Attention Computation:} For each candidate neighbor $j\in\mathcal{N}(i)$ of pixel $i$, compute $\alpha_{ij}^{(l)}$. Specifically, perform XNOR matching on $\mathbf{q}_i^{(l)}$ and $\mathbf{k}_j^{(l)}$ to obtain similarity $s_{ij}$, then output Boolean attention indicator $\alpha_{ij}^{(l)}\in\{0,1\}$ through comparison circuit. This process can be completed in highly parallel fashion in hardware, as XNOR computations for different neighbors are independent and only involve bit operations.

\paragraph{Attention-Gated Diffusion:} Perform bitwise AND on all neighbors $j$'s current state vectors $\mathbf{x}_j^{(l)}$ with corresponding attention $\alpha_{ij}^{(l)}$ to obtain gated neighbor state $\mathbf{x}'_{j} = \alpha_{ij}^{(l)} \wedge \mathbf{x}_j^{(l)}$ (if $\alpha_{ij}^{(l)}=0$ then this neighbor contributes all-zero vector). Then collect all gated neighbors of $i$ $\{\mathbf{x}'_j: j\in\mathcal{N}(i)\}$, together with own state $\mathbf{x}_i^{(l)}$ as input to subsequent logic kernel.

\paragraph{Logic Kernel Computation (Reaction):} Send pixel $i$'s own state and gated neighbor states into gate-level logic kernel $f_{\theta}$. Here all pixels share the same set of logic circuit structure and parameters $\theta$ (can also be extended to different pixels/layers having different parameters, but we assume sharing to reduce parameters). $f_{\theta}$ performs combinational logic computation on inputs, outputting pixel $i$'s new state $\tilde{\mathbf{x}}_i^{(l)}$ at this layer. Since $f_{\theta}$ is bitwise operation, for $m$-bit input it will produce $m$-bit output.

\paragraph{Boolean Activation and Normalization (Optional):} In some designs, we can introduce some Boolean-equivalent "activation" or "normalization" operations after logic kernel output to stabilize training. For example, each bit of output vector can be decided to keep or flip through a trainable threshold function (similar to majority voting logic) to control signal amplitude; or add simple error correction logic to avoid instability caused by too many neighbor signals. Commonly seen LayerNorm in Transformers does not directly apply in Boolean environments, but modules like "bitwise balancers" can be designed for logic adjustment. For theoretical simplicity, this paper does not specifically introduce this module.

\paragraph{Residual Connection and Iteration:} We can choose to directly use $\tilde{\mathbf{x}}_i^{(l)}$ as layer $l$ output, or add residual connection-like mechanisms, such as combining $\tilde{\mathbf{x}}_i^{(l)}$ with input $\mathbf{x}_i^{(l)}$ through logic OR (XOR) to obtain $\mathbf{x}_i^{(l+1)}$. Logic implementation of residuals can be diverse, such as using a one-bit switch to control whether to keep certain bits of old state or adopt new computed values, thereby alleviating degradation of deep logic networks. Finally, layer $l$ output $\mathbf{x}_i^{(l+1)}$ will be supplied to the next layer for continued processing.

After $L$ layers of stacked operations, we obtain output Boolean field $\mathbf{X}(L)$. If the task requires a Boolean image (e.g., segmented binary map), then $\mathbf{X}(L)$ can directly serve as the result. If the task requires real-valued output, such as classification probabilities or regression values, we can attach a small interpretable readout module on top of final Boolean state, such as counting how many pixels are 1 or how many times specific patterns appear, then obtaining corresponding output through lookup table. Since our focus is on image-to-image Boolean mapping process, we do not expand on readout modules here.

It's worth mentioning that this architecture has correspondences with classic Transformers: Boolean attention mechanism corresponds to Transformer's multi-head self-attention layer, except each "head" here uses XNOR to measure correlation and takes hard gating; logic kernel corresponds to feedforward network in Transformer (but we implement nonlinear transformation with logic gates); residual and layer-by-layer stacking make it a deep network. Different from Transformer is that our entire pathway doesn't involve numerical multiplication and addition, but is entirely composed of Boolean operations. This means in digital circuits or low-power hardware, this model's execution efficiency will be extremely high, with each operation unit occupying minimal hardware resources. Meanwhile, due to determinism of logic operations, model behavior is easy to analyze and verify, more transparent compared to continuous models.

\section{Trainable Parameters and Forward Propagation}
\label{sec:training}

This section discusses details of model parameterization, continuous relaxation techniques during training, and computational characteristics of forward propagation.

\subsection{Overview of Trainable Parameters}

Our model contains the following main learnable parameters:

\paragraph{Logic Kernel Parameters $\theta$:} Determine functional configuration of each gate in logic kernel circuits. For example, taking two-input gates as basic units, each gate has $g$ types to choose from ($g=16$ covers all two-input Boolean functions), then each gate corresponds to a $g$-dimensional parameter vector, gradually approaching a certain one-hot gate type during training. The set of all logic kernel gate parameters constitutes $\theta$. If different layers or different pixels use independent logic kernels, there will also be corresponding multiple sets of parameters, depending on model variants.

\paragraph{Query/Key Generation Parameters:} Used to compute $\mathbf{q}$ and $\mathbf{k}$ from pixel current state. If this function is determined by gate-level circuits or lookup tables, then corresponding Boolean function configurations are parameters. For instance, if using a $k\times d$ Boolean matrix to project $m$-bit state to $d$-bit query (similar to linear transformation, but implemented on $\{0,1\}$ or $\{\text{AND/OR/XOR}\}$ combinations), then this matrix can be trained. Generally, we can have parameter $\phi$ representing query generation function and $\psi$ representing key generation function configuration.

\paragraph{Attention Threshold/Decision Parameters:} If comparison threshold $\tau$ or decision logic in attention module is not fixed, it can also be included in training. For example, a simple trainable scheme is to introduce a learnable bias $b$ on similarity count $s_{ij}$ basis, deciding $\alpha_{ij}$ by judging $s_{ij} + b \ge \tau$. $b$ adjusts during training, equivalent to bias factor in softmax attention mechanism. More generally, configuration of attention decision logic circuits can also be trainable, with parameters possibly included in $\phi,\psi$ or listed independently.

\paragraph{Position Encoding Parameters (Optional):} If we adopt fixed parity flipping encoding, there are no additional parameters. If wishing to learn position encoding patterns, we can give each possible relative displacement a $d$-bit code as parameter table. However, this will have too many parameters for large-size images, usually still better to use fixed design.

In summary, model parameter set $\Theta = \{\theta, \phi, \psi, b, \ldots\}$, where $\theta$ dimension is proportional to logic kernel circuit complexity, $\phi,\psi$ relate to $m,d$ dimensions.

\subsection{Forward Propagation and Continuous Relaxation}

Forward propagation strictly follows logic steps described in the previous section. When the model is fully discretized, each pixel's update at each layer is pure Boolean circuit computation with deterministic output. However, directly using discrete operations during training causes gradients to be zero or unstable. To this end, we introduce continuous relaxation and approximation during training, making the network exhibit (sub-)continuous response to parameter changes.

The specific method is to map Boolean values 0 and 1 to real-number scalars, such as mapping 0 to 0.0, 1 to 1.0 (or -1 and +1 mapping is also feasible). Then replace main discrete operations with differentiable approximations:

\paragraph{Basic Logic Operator Relaxation:} Real-valued logic can be used as replacement. For example, using real multiplication to approximate AND (because $1\cdot 1=1$, other cases $<1$), using bitwise maximum to approximate OR (because $\max(0,x)$ can analogize OR), etc. Or neural networks can be used to approximate each truth table. However, a more general method is to adopt mixed gate representation for each gate, using softmax weights to combine multiple gate outputs. Petersen et al.~\cite{petersen2022deep} use this approach to implement continuous weighted output of all possible gates at each logic unit, equivalent to "expected output". This output is continuously differentiable with respect to gate weights.

\paragraph{XNOR Attention Relaxation:} For XNOR matching degree, we can directly use dot product or cosine similarity for approximation. Another clever method is utilizing arithmetic equivalent forms of logic operations: XOR can be expressed as $x \oplus y = x + y - 2xy$ (in 0/1 representation), then XNOR is $1 - (x \oplus y)$. Extending this to vectors,

\begin{equation}
\tilde{s}_{ij} = \frac{1}{d}\sum_{m}(1 - (q_{i,m}+k_{j,m}-2q_{i,m}k_{j,m})) = 1 - \frac{2}{d}\sum_m q_{i,m}k_{j,m} + C
\end{equation}

where $C$ is a constant related to Hamming weight of $q,k$ but independent of matching. Ignoring $C$, $\tilde{s}_{ij}$ is linearly related to $q_i^\top k_j$. This suggests we can use dot product or correlation to replace counting. (Note: when $q,k$ take $\{-1,+1\}$ representation, XNOR count is proportional to dot product) During training, adopt this \textbf{continuous similarity}, and convert it to 0-1 probability through Sigmoid, used to represent $\alpha_{ij}$. For example, $\Pr(\alpha_{ij}=1) = \sigma(\lambda (s_{ij}-\tau))$, where $\lambda$ is sharpness coefficient. Forward takes $\sigma$ value as soft attention weight participating in computation, backward gradient can propagate. After training, harden to threshold comparison.

\paragraph{Other Discrete Decision Relaxation:} If needing to assign 1 to the largest few neighbors of $s_{ij}$, softmax can be used to form differentiable approximation similar to top-$k$ selection. For logic kernel output binarization, real values can be temporarily retained and stacked, or the famous Straight-Through Estimator (STE) strategy can be used: forward uses Heaviside step to obtain Boolean output, backward ignores step gradient, directly passing gradient to real value before step. This is a commonly used trick in binary network training.

Through the above relaxation, during forward propagation our model actually becomes a \textbf{hybrid-valued network}: intermediate signals are continuous probabilities or real numbers, but network structure still strictly corresponds to some Boolean circuit. When training converges, we expect each relaxed unit to approach discrete state (e.g., softmax becomes one-hot, Sigmoid approaches 0 or 1, etc.). At this time, we can map the entire network back to Boolean implementation for inference. This "soft training, hard execution" paradigm has been proven practically feasible in logic networks~\cite{petersen2022deep}.

\subsection{Training Process}

We adopt conventional supervised learning methods to train model parameters. Given training dataset $\{(\mathbf{X}_{\text{in}}^{(n)}, \mathbf{Y}_{\text{out}}^{(n)})\}_{n=1}^N$, where $\mathbf{Y}_{\text{out}}^{(n)}$ is expected output (can be Boolean image or other labels), define loss function:

\begin{equation}
\mathcal{L}(\Theta) = \frac{1}{N}\sum_n \text{loss}(F(\mathbf{X}_{\text{in}}^{(n)};\Theta), \mathbf{Y}_{\text{out}}^{(n)})
\end{equation}

where $F$ represents our model's forward mapping. During forward, use continuous relaxation version $F_{\text{soft}}$ to compute output, then calculate loss based on $\mathbf{Y}_{\text{out}}$ (such as pixel cross-entropy, IoU loss, etc.). Iteratively update $\Theta$ using gradient descent or Adam optimization until convergence. During training, can progressively \textbf{tighten relaxation} (e.g., increase $\lambda$ or decrease temperature) to gradually binarize output, smoothly transitioning to discrete rules. To encourage interpretable logic, we can also add \textbf{regularization terms}, such as penalizing entropy of logic gate distribution, jitter of attention gates, etc., making each unit make clear 0/1 decisions as much as possible.

\subsection{Inference and Discretization}

After training completes, we obtain a set of approximately discrete parameters $\Theta^*$. We perform discrete mapping on continuous parts: taking $\arg\max$ of each logic gate's parameter $\mathbf{p}$ for corresponding gate type, taking each attention threshold closest to making Sigmoid output 0/1 boundary, etc. This constructs a deterministic Boolean network implementation $F_{\text{hard}}$. During inference, for any new input Boolean image, layer-by-layer execution of XNOR, AND and other logic operations can obtain output without any real-number computation. Since the entire model is essentially \textbf{combinational logic circuit}, inference latency is equivalent to circuit's logic depth and can be massively parallelized, with hardware performance far exceeding equivalent floating-point networks. This property is very advantageous for deployment on resource-constrained devices (such as FPGAs, microcontrollers).

\section{Theoretical Analysis and Properties}
\label{sec:theory}

This section analyzes properties of our proposed Boolean geometric attention network from perspectives of expressive power, computational complexity, and interpretability.

\subsection{Expressive Power and Equivalence}

Gate-level Boolean networks have a different representation paradigm from traditional networks. Theoretically, a network containing enough logic gates can represent any Boolean function (because logic gate sets like \{AND, OR, NOT\} are complete). In our design, since each layer's logic kernel processes local neighborhoods, our model is homologous to \textbf{locally ruled cellular automata}. If neighborhoods are large enough, our network can approximately simulate arbitrary Boolean mappings on space, including effects of traditional convolutional neural networks. For example, a $3\times3$ convolution kernel applied to binary images is equivalent to some fixed 9-input Boolean function, and our logic kernel can completely learn equivalent logic (such as performing weighted sum thresholding, equivalent to approximating convolution plus ReLU). In fact, existing research has shown that circuits equivalent to linear threshold units can be constructed by combining logic gates, thereby reproducing perceptron models. Additionally, through multi-layer stacking and attention, global dependencies can also be established, similar to Transformer having global modeling capabilities. Therefore, we have reason to believe that as long as network scale (number of layers, number of logic gates) is sufficient, our model's \textbf{approximation capability} in Boolean domain is no less than real-valued networks' approximation capability in continuous domain. In particular, if discretizing real-valued network weights to Boolean decisions, our model can simulate by learning corresponding logic rules.

\subsection{Dynamic Behavior and Stability}

Since the model is essentially a \textbf{discrete dynamical system}, oscillation, convergence, or complex cycles may occur in closed-loop use. Like classical cellular automata, finite-sized Boolean network state space is finite, so evolution must enter cycles or fixed points. However, through well-designed logic kernels, we can control this dynamic. For example, rules that \textbf{suppress oscillation} can be learned, making the network converge to stable patterns within finite steps (such as making boundaries converge stably in image segmentation). Addition of attention mechanism provides \textbf{context-sensitive} regulation, hopefully reducing unnecessary oscillation: because when certain patterns form, attention may selectively turn off influence of some interfering neighbors, thereby maintaining stable state. Of course, if convergence is not desired, \textbf{oscillators} or specific periodic cycles can also be designed using logic rules (similar to oscillating structures in CA). Overall, our model can exhibit rich dynamics, both learning stable solutions through training and presenting periodic behavior as needed. This malleable dynamic property provides flexibility for tasks like image generation and visualization.

\subsection{Computational Complexity and Hardware Efficiency}

On classical computers, although Boolean operations have low single-operation overhead, higher bit-width parallelism may be needed to match floating-point operation volume. However, our model's advantages lie in being \textbf{highly parallelizable} and having \textbf{gate-level low latency}. XNOR and AND operations in each layer can be viewed as bit-level matrix multiplication, which can be completed at bit-level in parallel in hardware implementation. For example, a 32-bit register can simultaneously execute XNOR for 32 pixel pairs, and modern CPU SIMD instructions or dedicated circuits can even complete thousands of bits of parallel XNOR within clock cycles. Logic gates have extremely low area and energy consumption in digital circuits, for example, a 1-bit XNOR gate requires only one FPGA logic unit, while a 32-bit floating-point multiplier requires about 200 units. Therefore, on dedicated chips, our model has potentially huge speed and energy efficiency advantages. Meanwhile, since output is discrete values, complex quantization processes are unnecessary, avoiding quantization errors. In software implementation, if using bit operation optimization, speed orders of magnitude faster than floating-point operations can also be achieved. When model scale increases, its computational load mainly grows linearly with neighborhood size; compared to Transformers requiring fully connected computation, our attention is typically limited to local areas, thus having lower \textbf{time complexity} (per pixel $O(|\mathcal{N}|d)$, usually $|\mathcal{N}|d$ much smaller than image pixel count). Overall, our model promises near real-time performance on large-scale image tasks, especially suitable for resource-constrained scenarios.

\subsection{Interpretability and Rule Extraction}

Since the model ultimately discretizes to logic gate connections, we can \textbf{completely explain} the model's decision process. For instance, in classification tasks, we can trace layer-by-layer how pixels affect final output through logic operations and extract "If-then"-like rules: such as "if region A has certain shape pattern and region B has no noise, then output is 1", which can be precisely expressed in logic gate networks. While attention introduction dynamically changes connections, attention itself is also determined by Boolean conditions, which can be further parsed into rules (such as "ignore neighborhood Y when pattern X appears"). Compared to traditional neural networks needing heuristic methods for explanation (and explanations being incomplete), our model can provide \textbf{intrinsic interpretability}. This is very beneficial for high-security or decision-justification-requiring applications. Additionally, we can perform \textbf{formal verification} on logic networks: using Boolean algebra or SAT solvers to verify whether certain properties (such as robustness, invariance) hold, which is almost impossible to directly apply to large floating-point networks. Therefore, in domains requiring strict verification of AI behavior (autonomous driving, medical diagnosis, etc.), this model has unique advantages.

\subsection{Relationship with Existing Models}

Our model can be viewed as combining elements of \textbf{cellular automata, graph attention networks, and Transformers}. From cellular automata perspective, we provide learnable rules enabling them to handle complex tasks; from graph attention perspective, we use logic to replace linear algebra, achieving discrete attention message passing; from Transformer perspective, we introduce multi-head attention and position encoding concepts, just with different implementation domains. An interesting extreme case is turning off attention (always $\alpha_{ij}=1$) and restricting logic kernel to linear threshold functions, then our model degenerates to classical CNN/GCN (equivalent to neighborhood weighted sum plus threshold activation). Another opposite case is restricting neighborhood to global (adjacent to all pixels) and having logic kernel only aggregate from attention-selected pixels, then model becomes similar to standard Transformer structure. Therefore, gate-level Boolean evolutionary networks cover a wide range in design space from local CA to global Transformer. Through training, it can find the optimal point for specific tasks in this space.

Of course, current model also has some limitations. First, complete Boolean discretization may make gradient optimization difficult; although we use relaxation techniques, training process may still be more time-consuming than standard networks, requiring careful adjustment of relaxation strategies. Second, due to using logic gates, model may be more sensitive to random noise (single bit flip may change output), requiring consideration of robustness in design (such as introducing redundancy codes, error correction logic). Third, model needs pre-quantization when processing continuous-valued inputs, which may cause information loss, though this can be partially alleviated by increasing bit dimensions.

\section{Potential Applications and Discussion}
\label{sec:applications}

\subsection{High-Speed Low-Power Image Processing}

The most direct advantage of gate-level Boolean networks lies in their hardware friendliness, suitable for implementing high-throughput computation in FPGAs, ASICs, or even memory arrays. Therefore, it has potential applications in image tasks requiring real-time processing. For example, in embedded vision systems, this model can execute frontend object detection, binary segmentation, etc., sinking computation to sensors or edge devices. Since inference only requires Boolean operations, power consumption is extremely low, extending battery life or reducing cooling requirements. In some latency-sensitive scenarios (such as autonomous driving perception systems, industrial defect detection), our model promises nanosecond-level response in hardware circuit form, which traditional deep networks find difficult to achieve.

\subsection{Interpretable and Verifiable AI}

In high-risk domains such as medical diagnosis and financial decision-making, people hope model decision processes are transparent and auditable. Our model provides possibility of representing decisions in forms close to human logic. For example, in medical image analysis, models can learn rules like "if tumor edge presents certain Boolean combination features then judge as malignant", helping doctors understand discrimination basis. Additionally, through logic refinement, we can verify whether model behavior under certain input patterns conforms to priors. For example, in autonomous driving vision, can verify whether safety rules like "if front vehicle distance too close and speed not reduced, then brake signal must be true" are strictly followed by model. This formal verification relies on model's logic circuit representation, can use SAT checkers to exhaustively verify all possible inputs, which continuous models cannot do. Therefore, our network is very suitable for applications in \textbf{scenarios requiring high reliability}.

\subsection{Image Generation and Complex Pattern Evolution}

Thanks to reaction-diffusion dynamics, this model also has uses in generating artistic patterns, simulated ecosystems, etc. Reaction-diffusion systems are widely used to simulate fish patterns, zebra stripes, and other natural patterns; through training, our logic network can learn Boolean rules producing similar complex patterns. For example, letting network evolve from random seeds to certain stable spot or stripe patterns, similar to learning a \textbf{pattern-generating cellular automaton}. Compared to continuous generative models, Boolean evolution generation has unique style, and each evolution step is interpretable as action of specific rules. Artists and designers can adjust logic rules to produce different effects, even directly edit logic circuits for controllable generation. This provides new pathways for \textbf{interpretable generative art}.

\subsection{Bridge Fusing Symbolic AI and Perception}

Our model naturally combines \textbf{symbolic logic} (through gate circuits) with \textbf{perception tasks}. On one hand, it can process pixel-level large-scale data; on the other hand, internal representation is symbolic. This characteristic gives it potential as a bridge connecting deep learning and symbolic AI. For example, Boolean images output by this model can be further input to symbolic reasoning modules (such as expert systems or knowledge bases), so perceived entities (represented in Boolean, like detecting certain feature=1) can seamlessly enter logic reasoning process. Conversely, symbolic constraints can also be translated into structural priors of this model (such as certain regions must satisfy specific logic relations to output true). Compared to completely black-box neural networks, our architecture more easily fuses knowledge and rules, thus having unique appeal in applications requiring domain knowledge combination (such as robot vision, industrial process control).

\subsection{Analogy with Biological and Physical Systems}

Since the model evolves with 0/1 states and local rules, it has similarities with discrete dynamical systems like biological neuron excitation/inhibition (on/off) and gene regulatory networks. We can abstract some biological pattern formation mechanisms into Boolean rules (such as activating certain genes when certain signals are high during cell division) and simulate their spatial behavior through this model. This has potential significance for \textbf{systems biology} research: through learning data-driven Boolean models, exploring simple rules of complex systems. Similarly, physical self-organized criticality phenomena can also be discretized and attempted to reproduce or predict using our model.

\subsection{Discussion}

Although this model demonstrates many advantages, there are also issues requiring further discussion. First, \textbf{model scale and training difficulty}: dramatic increase in logic gate numbers will increase training difficulty, because parameter space is discrete combinatorial and loss landscape may be very non-smooth. Whether current continuous relaxation techniques are effective, whether need to fuse evolutionary algorithms and other global optimization methods, are topics worth researching. Second, \textbf{how to design efficient Boolean attention}: our attention currently is hard selection, though concise but may not have as fine regulation capability as softmax weights. Future can explore multi-bit attention or hierarchical attention, letting attention intensity have more levels. Furthermore, \textbf{robustness and generalization}: logic networks sometimes overfit training distribution, lacking smooth response to tiny input changes. Can consider combining fuzzy logic or probabilistic logic, making model have certain fuzzy weights between 0 and 1, thereby improving generalization. Finally, \textbf{combination with continuous networks}: our framework doesn't necessarily have to completely replace CNNs or Transformers, can completely be embedded as modules into large networks, achieving hybrid models. For example, using continuous networks to extract features, then having Boolean modules make decisions, to simultaneously utilize advantages of both. These will be further explored in future work.

\section{Conclusion and Future Work}
\label{sec:conclusion}

This paper proposes a novel theoretical framework—\textbf{Geometric Attention Neural Networks Based on Gate-Level Boolean Evolution}—organically fusing image processing with logic computation, neighborhood diffusion, and attention mechanisms. By modeling image pixels as Boolean variables evolving on geometric manifolds, we implement a Transformer-like deep network architecture with all computational units composed of Boolean logic gates. We detailed model components including trainable reaction-diffusion logic kernels, XNOR similarity-based Boolean attention mechanisms, and Boolean rotary position encoding. Theoretical analysis shows this model has good expressive power and hardware efficiency, supporting complex spatial information processing while providing high interpretability.

This research provides new ideas for combining logic circuits with deep learning. In future work, we plan to verify this model's performance on several practical tasks, such as image segmentation, pattern generation, and image classification, comparing with equivalent real-valued models to evaluate comprehensive advantages in accuracy, speed, and energy consumption. Additionally, we will explore techniques to improve model trainability, including smoother continuous relaxation strategies, mixed integer optimization, and heuristic-based structure search, helping automatically discover high-performance logic network structures. We also consider applying this model to \textbf{neural architecture search} domains: due to its discrete composable characteristics, perhaps can be used to efficiently search networks meeting specific hardware constraints. Finally, we have interest in this model's interdisciplinary applications, such as using logic attention networks to simulate cellular automata and chemical reaction-diffusion processes, hoping to deepen understanding of complex system regularities from AI perspective.

In summary, \textbf{Gate-Level Boolean Geometric Attention Neural Networks} open a brand new path for executing neural computation in logical ways. We believe that as research progresses, this direction will demonstrate huge potential in realizing efficient, transparent, and verifiable intelligent systems, providing beneficial supplements and references for next-generation artificial intelligence models.

\section*{Acknowledgments}

We thank the reviewers for their patience in reading this paper draft. Comments and suggestions are welcome.

\bibliographystyle{plain}

\end{document}